# Multi-body dynamic evolution sequence-assisted PSO for interval analysis

Xuanlong Wu[1,*], Peng Zhong[1], Weihao Lin[1]

[1] State Key Laboratory of Structural Analysis, Optimization and CAE Software for Industrial Equipment, School of Mechanics and Aerospace Engineering, Dalian University of Technology, Dalian 116024, PR China

\* Correspondence: wuxuanlong7895@mail.dlut.edu.cn

**Abstract:** When the exact probability distribution of input conditions cannot be obtained in practical engineering problems, interval analysis methods are often used to analyze the upper and lower bounds of output responses. Essentially, this can be regarded as an optimization problem, solvable by optimization algorithms. This paper proposes a novel interval analysis method, i.e., multi-body dynamic evolution sequence-assisted PSO (abbreviated as DES-PSO), which combines a dynamical evolutionary sequence with the heterogeneous comprehensive learning particle swarm optimization algorithm (HCLPSO). By introducing the dynamical evolutionary sequence instead of the random sequence, the proposed method addresses the difficulty HCLPSO faces in covering the search space, making it suitable for interval analysis problems. To verify the accuracy and efficiency of the proposed DES-PSO method, this paper solves two case studies using both the DES-PSO and HCLPSO methods. The first case study employs an optimization algorithm to solve the solution domain of a linear interval equation system, and the second case study analyzes the collision and heat conduction of a smartwatch using an optimization method. The results of the case studies demonstrate that DES-PSO can significantly improve the computational speed of interval analysis while ensuring accuracy, providing a new approach to solving complex interval analysis problems.

1. Introduction

Uncertainty is pervasive in practical engineering problems, such as material properties [1-6], manufacturing errors [7-11], random loads [12-16], service life [17-19], and trajectory tracking [20]. The presence of uncertainty in measurement data and mathematical models is inevitable [21-23], and can significantly affect structural responses, necessitating evaluation [24]. Even slight variations in numerous uncertainty factors can lead to considerable deviations in structural or system responses. Coleman and Steele [25] pointed out that in an experiment where all measurements have only 1% uncertainty, the uncertainty of the experimental results can exceed 50%, implying that underestimating or incorrectly estimating uncertainty can severely impact the safety of engineering structures, potentially leading to catastrophic outcomes.

Therefore, it is crucial to consider the impact of uncertainty factors in structural analysis [26, 27].

In practical engineering problems, due to limitations in experimental conditions and costs, some parameters cannot be accurately modeled with probability distributions due to insufficient experimental samples, leaving only their variation intervals available [28, 29]. Given the interval uncertainty of input parameters, structural responses also exhibit interval uncertainty, making it vital to evaluate the interval uncertainty of responses. Failure to correctly estimate the variation range of responses could lead to erroneous judgments about the safety of materials or structures, posing serious safety risks [30, 31]. Hence, designing precise and efficient interval analysis methods for structural responses is of significant engineering importance.

Commonly used interval analysis methods include the Monte Carlo method (MCM), interval perturbation method, Chebyshev interval method, and vertex method. MCM provides stable solutions. With sufficient sample data, it can ensure accuracy [32]. However, in large-scale engineering problems, the computational costs will be greatly increased. Interval perturbation is a straightforward simplification method. Qiu et al. [33] solved displacement responses of uncertain parameters under small interval widths using the perturbation method, which later developed into the interval perturbation method [34]. However, the solution of this method may not converge under large interval dispersion. Xia et al. [35, 36] proposed a sub-interval perturbation method to calculate the response domain of interval acoustic coupling systems. However, as the number of interval parameters and sub-intervals increases, the computational burden grows significantly. Wu et al. [37, 38] combined Chebyshev polynomial series expansion to address interval uncertainty, but the computational complexity of Chebyshev polynomial series increases exponentially with the number of parameters, leading to dimensionality issues. Qiu et al. [39, 40] also combined finite element analysis with non-random convex models to propose the vertex method, suitable for solving exact boundaries of linear interval equations. However, this method is limited to monotonic cases, and its computational cost increases exponentially with interval parameters, making it impractical for large structural problems.

The essence of interval analysis lies in calculating the upper and lower bounds of structural responses based on the interval range of input parameters, which can be viewed as finding the maximum and minimum values of structural responses, thus forming two optimization problems. Therefore, interval analysis problems are often solved using optimization algorithms [41, 42]. Common optimization problem-solving methods can be divided into two categories: traditional gradient-based algorithms, characterized by fast convergence and high computational efficiency, but prone to getting trapped in local optima when dealing with multi-modal problems, and meta-heuristic intelligent optimization algorithms, which are less efficient than traditional gradient methods but have a strong ability to escape local optima and are considered effective for finding global optima. Many scholars have attempted to apply intelligent optimization algorithms to interval analysis. Feng et al. [43] applied the Bayesian global optimization algorithm to the interval estimation of safety factors in deterministic slope stability. Cheng et al. [44] proposed an improved Pelican algorithm and used it for interval analysis of free vibration responses of 3D pyramidal truss core sandwich panels. Ta et al. [45] proposed a new interval particle swarm optimization algorithm based on particle swarm optimization, applying it to interval analysis of vehicle body vibration and optimization of aerial camera stability. Sharma and Jabeen [46] combined interval methods with the artificial bee colony algorithm to solve interval analysis problems in real-world constrained engineering.

The degree of solving interval analysis problems based on optimization algorithms is closely related to the computational performance of the optimization algorithms used. In recent years, more powerful optimization algorithms have emerged, such as the adaptive grey wolf optimizer (AGWO) [47], improved Harris Hawks optimization (IHHO) [48], and hybrid pathfinder algorithm (HPFA)[49]. Among them, the heterogeneous comprehensive learning particle swarm optimization algorithm (HCLPSO) [50] proposed by Lynn and Suganthan balances the global and local search capabilities of the particle swarm optimization algorithm and is recognized as one of the most outstanding variants of the famous PSO algorithm. It has been successfully applied in areas such as aircraft control [51], permanent magnet motor parameter estimation [52], and wave spectrum fitting [53]. However, no research has yet applied this method to interval uncertainty analysis.

Many existing interval analysis problems are often multi-modal, making the application prospects of intelligent optimization algorithms in interval uncertainty analysis highly anticipated. HCLPSO, with its outstanding global and local search capabilities, is expected to provide important support for constructing efficient and accurate interval analysis algorithms. However, HCLPSO searches for optima by generating random point sets, which leads to uneven point distribution, making it difficult to effectively cover the search space. This paper introduces HCLPSO into interval uncertainty analysis and enhances its search space coverage by using the low discrepancy sequence generated through dynamic evolution sequence (DES) [54], resulting in a new interval uncertainty analysis method, DES-PSO, which significantly improves computational efficiency while maintaining the accuracy of HCLPSO.

The remainder of this paper is structured as follows: Section 2 briefly introduces interval problems. Section 3 reviews the algorithmic principles of HCLPSO. Section 4 presents a new interval uncertainty analysis method, DES-PSO, which combines enhanced exploration and exploitation of heterogeneous particle swarm optimization with sequence generated by dynamic evolution algorithm. Section 5 validates the effectiveness of DES-PSO in handling interval uncertainty problems through the solution of an interval equation system and an engineering case study involving the design of a smartwatch case. The conclusion is given in Section 6.

## 2. Brief introduction to interval problems

When dealing with practical engineering problems, it is often difficult to obtain a large amount of sample information, which makes it impossible to describe the uncertainty of random variables $X$ with an exact probability distribution function. To address this issue, many scholars typically use the method of interval models to solve [13], which can greatly reduce time and computational costs. When characterizing its uncertainty using interval models, the uncertain variable $X$ is referred to as an interval variable and can be represented as

$$X = [x_1, x_2, \cdots, x_D], a_i \leq x_i \leq b_i, \quad (1)$$

where the subscript $i \in 1, 2, \cdots, D$, $D$ denotes the number of variables; $a_i$ and $b_i$ represent the lower and upper bounds of the interval for the variable $x_i$, respectively.

Based on the interval variable $X$, a general form of interval analysis model can be established as

$$Y = f(X), \quad (2)$$

where $f(\cdot)$ is the structural response function, and $Y$ is the output uncertainty response.

Since the interval variable $X$ is used as input, the output of the response function will vary within an interval rather than being a definite value, which is $Y = [y_1, y_2, \cdots, y_M]$, $y_i \in [y_{i,\min}, y_{i,\max}]$. Each $y_i$ corresponds to a structural response function $f_i(\cdot)$. The purpose of interval uncertainty analysis is to determine the response interval of the response function, i.e., to ascertain the boundaries of the output response. Therefore, the problem of interval uncertainty analysis for a structure can be transformed into the following two optimization problems:

$$\begin{cases} y_{i,\min} = \min(f_i(X)) \\ y_{i,\max} = \max(f_i(X)) \end{cases}. \tag{3}$$

## 3. HCLPSO

Particle swarm optimization (PSO) is a population-driven evolutionary strategy [55, 56]. In this strategy, each potential solution is symbolically represented as a soaring bird, which optimizes its position in the solution space based on its own flight experience and the experiences of other members of the population. Under the framework of Heterogeneous Comprehensive Learning Particle Swarm Optimization (HCLPSO) [50] to achieve an in-depth exploration of the global search and a fine exploitation of local search, the entire population is divided into two specialized subpopulations, each responsible for exploration and exploitation tasks.

To simplify our subsequent analysis process, HCLPSO is presented in a matrix-vector form, and the population size of the problem is defined as $N$, the maximum number of iterations as $G$, the objective function as $f$, and the feasible solution as $x = (x_1, x_2, \cdots, x_D)^T \in R^{D \times 1}$, where $x_i \in [a_i, b_i]$. Let

$$a = (a_1, a_2, \cdots, a_D)^T, \quad b = (b_1, b_2, \cdots, b_D)^T. \tag{4}$$

Let the population be represented as

$$X_g = [x_{g,1}, x_{g,2}, \cdots, x_{g,N}], \tag{5}$$

where the subscript $g$ represents the $g$-th iteration step. The initial population in HCLPSO is

$$X_0 = a \otimes I_N + \varepsilon_0 \circ (b \otimes 1 - a \otimes 1), \tag{6}$$

where $\otimes$ and $\circ$ denote the Kronecker product and the Hadamard product, respectively, $I_N$ is a vector of size $1 \times N$, with each element being 1, and $\varepsilon_0$ is a matrix of random numbers uniformly distributed between 0 and 1, with a matrix size of $D \times N$.

When iterating the population of HCLPSO, the updating method is as follows:

$$x_{g+1,i} = x_{g,i} + v_{g+1,i}, \quad i = 1, 2, \cdots, N, \tag{7}$$

where $v_{g+1,i}$ is the velocity of the $i$-th particle. The population of HCLPSO is divided into exploration and exploitation subpopulations, with the sizes of the two types of subpopulations being $N_1$ and $N_2$, respectively, and their velocity updating formulas are also different. The expression for the velocity updating formula of the exploration subpopulation is

$$v_{g+1,i} = w_g v_{g,i} + k_g \varepsilon_{g,1,i} \circ \left( p_{g,i} - x_{g,i} \right), \quad 1 \leq i \leq N_1, \tag{8}$$

where $\varepsilon_{g,1,i}$ is a vector of random numbers uniformly distributed between 0 and 1, with a vector size of $D \times 1$, and $p_{g,i}$ is a random comprehensive learning vector, which enables the *i*-th particle to learn from the best experiences of all other particles. The relevant calculation formula for $p_{g,i}$ can be found in Ref. [50]. The expression for the velocity updating formula of the exploitation subpopulation is

$$v_{g+1,i} = w_g v_{g,i} + c_{g,1} \varepsilon_{g,2,i} \circ \left( p_{g,i} - x_{g,i} \right) + c_{g,2} \varepsilon_{g,3,i} \circ \left( x_{g,\text{best}} - x_{g,i} \right), \quad N_1 < i \leq N, \tag{9}$$

where $c_{g,1} = 2.5 - 2g/G$ and $c_{g,2} = 0.5 + 2g/G$ are used in HCLPSO, $\varepsilon_{g,2,i}$ and $\varepsilon_{g,3,i}$ are two vectors of random numbers uniformly distributed between 0 and 1, with both vector sizes being $D \times 1$, and $x_{g,\text{best}} = \arg\min\left( f\left( x_{g,i} \right) \right), \, 1 \leq i \leq N$.

HCLPSO possesses outstanding global and local search capabilities, which are expected to provide significant assistance in constructing efficient and precise interval analysis algorithms. However, HCLPSO searches for optimization by generating random sequences, which has the drawback of potentially failing to effectively cover the search space due to the uneven distribution of the sequences.

## 4. The proposed method

### 4.1. Dynamic evolution sequence

DES is a novel method for generating the low discrepancy sequence (LDS) recently proposed in Refs. [54, 57]. The concept of LDS originates from an in-depth observation of physical phenomena, where many multi-body systems in nature have static solutions with good uniformity. In the DES method, all particles in space are subject to gravitational forces acting between each other.

Assuming that all particles in space are within a hypercube $\Omega = [0, 1]^D$, where $D$ denotes the spatial dimension, which is also the number of variables, and the sample sequence is $X_{D,N} = \{x_1, \cdots, x_N\}$, with each sample point considered as a star of mass $m$, the coordinates of the *i*-th star are $x_i = (x_{i1}, \cdots, x_{iD})$, and there are interactive forces between points. The Lagrangian equations for these $N$ stars are

$$\begin{cases} S = \int_0^t L \mathrm{d}\tau \\ L = \dfrac{1}{2} m \sum_{i=1}^{N} \sum_{k=1}^{D} \dot{x}_{ik}^2 - G \left( \sum_{1 \le i < j \le N} \dfrac{1}{d_{q,ij}^p} \right)^{\frac{1}{p}} \end{cases}, \tag{10}$$

where $G = 1$ is the generalized gravitational constant; $q$ and $p$ are control parameters that affect the optimization performance of the algorithm, with specific values available in Ref. [54], and the expression for $d_{q,ij}$ is

$$d_{q,ij} = \sqrt{\sum_{k=1}^{D} \dfrac{\left|x_{ik} - x_{jk}\right|^2 \left(1 - \left|x_{ik} - x_{jk}\right|\right)^2}{\left[\left(1 - \left|x_{ik} - x_{jk}\right|\right)^q + \left|x_{ik} - x_{jk}\right|^q\right]^{\frac{2}{q}}}}. \tag{11}$$

According to the variational principle of Hamilton [58], it can be derived that

$$m\ddot{x}_{ik} + f_{ik} = 0, \ 1 \le i \le N, \ 1 \le k \le D, \tag{12}$$

in which,

$$\begin{cases} f_{ik} = -G \left( \sum_{1 \le i < j \le N} \dfrac{1}{d_{q,ij}^p} \right)^{\frac{1-p}{p}} \sum_{\substack{j=1 \\ j \ne i}}^{N} \dfrac{a_{ijk}}{d_{q,ij}^{p+2}} \\ a_{ijk} = \dfrac{\mathrm{sgn}\left(x_{ik} - x_{jk}\right) \Delta_{ijk} \left(1 - \Delta_{ijk}\right) \left[\left(1 - \Delta_{ijk}\right)^{q+1} - \Delta_{ijk}^{q+1}\right]}{\left[\left(1 - \Delta_{jik}\right)^q + \Delta_{ijk}^q\right]^{1+\frac{2}{q}}} \\ \Delta_{ijk} = \left|x_{ik} - x_{jk}\right| \end{cases}. \tag{13}$$

For all sample points to reach a stationary state, an artificial damping force is applied to each sample, thus Eq. (12) can be represented as the following common dynamical equation [59]:

$$m\ddot{x}_{ik} + c\dot{x}_{ik} + f_{ik} = 0, \ 1 \le i \le N, \ 1 \le k \le D, \tag{14}$$

where $c$ is the damping coefficient.

Eq. (14) is a nonlinear dynamical equation. This subsection uses a symplectic algorithm [58, 60] to discretize the equation: First, an initial sequence is given as $X^{(0)} = \left\{x_1^{(0)}, \cdots, x_N^{(0)}\right\}$, where $x_{ik}^{(0)}$ represents the $i$-th sample point in the $k$-th dimension at the initial iteration step. The basic iteration formula of the algorithm can be expressed as

$$\begin{cases} x_{ik}^{(1)} = x_{ik}^{(0)} - \dfrac{1}{2} m^{-1} f_{ik}^{(0)} \Delta t^2 \\ x_{ik}^{(g+1)} = \dfrac{A_1 x_{ik}^{g} - A_2 x_{ik}^{(g-1)} - f_{ik}^{(g)}}{A_0} \end{cases}, \tag{15}$$

in which

$$\begin{cases} f_{ik}^{(g)} = -\left( \sum_{1 \leq i < j \leq N} \frac{1}{\left(d_{q,ij}^{(g)}\right)^p} \right)^{\frac{1}{p}-1} \sum_{\substack{j=1 \\ j \neq i}}^{N} \frac{a_{ijk}^{(g)}}{\left(d_{q,ij}^{(g)}\right)^{p+2}} \\ A_0 = \frac{m}{\Delta t^2} + \frac{c}{2\Delta t}, \quad A_1 = \frac{2m}{\Delta t^2}, \quad A_2 = \frac{m}{\Delta t^2} - \frac{c}{2\Delta t}, \\ m = \frac{1+\kappa}{4} \frac{U\left(X^{(0)}\right)}{\left\|X^{(0)}\right\|_F^2} \Delta t^2, \quad c = \frac{U\left(X^{(0)}\right)}{\left\|X^{(0)}\right\|_F^2} \sqrt{\kappa} \Delta t \end{cases} \tag{16}$$

$\left\|X^{(0)}\right\|_F$ is the Frobenius norm; $\Delta t$ and $\kappa$ are also control parameters that affect the optimization performance of the algorithm, with specific values available in Ref. [54].

When all particles are in a stationary state, the generated sequence exhibits excellent uniformity, and this sequence has been proven to perform well in areas such as statistical problem-solving, intelligent optimization, experimental design, and physical model simulation [61-63].

*4.2. Dynamic evolution sequence improved HCLPSO (DES-PSO)*

From Eqs. (6), (8), and (9), it can be seen that the initial population and velocity updating formulas of HCLPSO are random sequences, which have a relatively poor effect in covering the search space compared to uniformly distributed populations. In order to improve the search efficiency and solution quality of HCLPSO, the DES sequence introduced in subsection 4.1 is applied to HCLPSO in this subsection.

In this subsection, the random sequences in the velocity updating formulas of HCLPSO are replaced with the LDS, and Eqs. (8) and (9) are rewritten as

$$v_{g+1,i} = w_g v_{g,i} + k_g P_i^{(1)} \circ \left(p_{g,i} - x_{g,i}\right), \quad 1 \leq i \leq N_1, \tag{17}$$

$$v_{g+1,i} = w_g v_{g,i} + c_{g,1} P_i^{(2)} \circ \left(p_{g,i} - x_{g,i}\right) + c_{g,2} \varepsilon_{g,3,i} \circ \left(x_{g,\text{best}} - x_{g,i}\right), \quad N_1 < i \leq N, \tag{18}$$

where $P^{(1)} = \left(P_1^{(1)}, \cdots, P_{N_1}^{(1)}\right)$ is a LDS of size $D \times N_1$, and $P^{(2)} = \left(P_1^{(2)}, \cdots, P_{N-N_1}^{(2)}\right)$ is a LDS of size $D \times (N - N_1)$.

Similarly, the random initial population of HCLPSO is replaced with an initial population generated by the LDS, i.e.,

$$X_0 = a \otimes 1_N + P^{(0)} \circ \left(b \otimes 1 - a \otimes 1\right), \tag{19}$$

where $P^{(0)}$ is a LDS of size $D \times N$.

This paper names the method of improving HCLPSO with dynamic evolution sequence as DES-PSO. Two numerical examples will be selected for comparison in the next section to verify the correctness and efficiency of DES-PSO.

## 5. Numerical examples

### 5.1. Linear interval equation systems

This example considers the solution of the following linear interval equation system, as shown in the following equation

$$\begin{bmatrix} [3,6] & [-3,1.5] \\ [-1.5,3] & [3,6] \end{bmatrix} \begin{Bmatrix} x_1 \\ x_2 \end{Bmatrix} = \begin{Bmatrix} [-4,4] \\ [-4,4] \end{Bmatrix}. \quad (20)$$

Reference [64] has studied this problem and plotted the solution domain, as shown in Figure 1. From Figure 1, it can be seen that the solution domain of this problem is non-convex, and the optimization problem at this time may have more than one extreme point. This naturally requires the use of a stochastic optimization algorithm for solving.

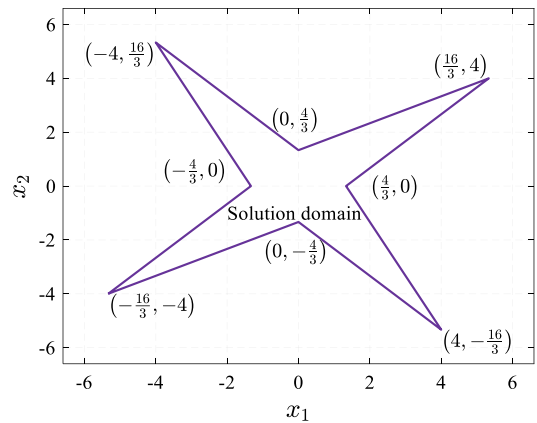

**Figure 1.** The solution domain of Eq. (20).

In this subsection, the DES-PSO proposed in subsection 4.2 and the original HCLPSO algorithm are used to solve this interval equation system simultaneously. The computational results show that both algorithms can obtain the exact upper and lower bounds, namely $[x_1] = [-16/3, 16/3]$ and $[x_2] = [-16/3, 16/3]$, which indicates that both the DES-PSO and the original HCLPSO algorithms have good optimization accuracy. To further compare the computational efficiency of the two algorithms, the number of convergence steps required for each algorithm to find the exact solution of the interval upper and lower bounds is counted, and the results are listed in Table 1.

**Table 1.** Comparison of convergence steps of two algorithms.

| Variable name | Interval lower bound | | Interval upper bound | |
|---|---|---|---|---|
| | DES-PSO | HCLPSO | DES-PSO | HCLPSO |
| $x_1$ | 2452 | 3371 | 2249 | 3211 |
| $x_2$ | 1904 | 3779 | 1379 | 3077 |

From Table 1, it can be seen that the convergence speed of DES-PSO is much faster than the convergence speed of the original HCLPSO. To better measure the degree of speed improvement, we introduce the relative speedup percentage as a measure of acceleration. The relative percentage can be expressed as (HCLPSO convergence steps - DES-PSO convergence steps) / DES-PSO convergence steps $\times 100\%$. When solving the lower bound of $x_1$, DES-PSO can achieve a relative speedup of 27.26%, and when solving the upper bound of $x_1$, DES-PSO can achieve a relative speedup of 30.93%. When solving the lower bound of $x_2$, DES-PSO can achieve a relative speedup of 49.61%, and when solving the upper bound of $x_2$, DES-PSO can achieve a relative speedup of 55.18%. Combining the results from Table 1, it can be concluded that our improved algorithm is more computationally efficient when dealing with interval problems.

*5.2. Uncertainty interval analysis of smartwatches*

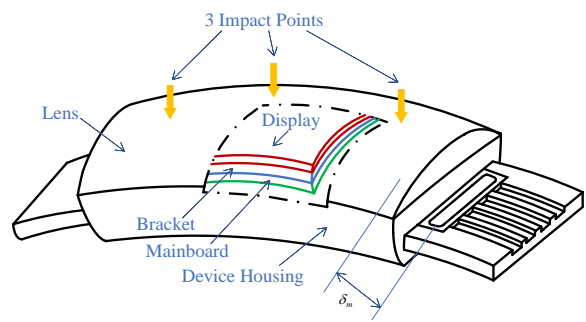

**Fig 2.** Schematic diagram of the smart watch.

Dynamic impact analysis [65-68] and heat conduction analysis [69-73] play significant roles in the field of engineering applications, especially for precision electronic devices [74]. Wearable electronic devices have a high degree of integration, during the design process, it is necessary to consider the requirements of mechanical, electrical, and thermal performance simultaneously. For example, it is necessary to ensure that the stress and the temperature of the embedded chips in the watch are within a reliable working range under impact conditions and high-temperature environments. This example uses the smartwatch model provided in Ref. [75], which is shown in Figure 2. This example studies the stress $\sigma_1^N$, $\sigma_2^N$, and $\sigma_3^N$ distribution intervals of the three different impact points shown in Figure 2, under impact conditions, and the stress $\sigma^H$ of the solder between the display and the mainboard, as well as the temperature $T_1$ and $T_2$ distribution intervals of the two embedded chips in the smartwatch under high-temperature environments.

The random variables and their distribution intervals in the watch are shown in Table 2. The interval random variables such as the smartwatch shell thickness, mainboard thickness, bracket thickness, and display thickness have a significant impact on the stress at the impact points and the chip temperature. Therefore, Ref. [75] used 65 finite element samples, considering the 10 interval random variables in Table 2, to construct the response surface functions for the stress and temperature of the smartwatch, as detailed in Table 3

Table 2. Interval parameters of random variables.

| Variable name | Random variable | Distribution interval |
|---|---|---|
| Shell thickness | $X_1$ (mm) | (0.91, 1.09) |
| Mainboard thickness | $X_2$ (mm) | (0.91, 1.09) |
| Bracket thickness | $X_3$ (mm) | (0.91, 1.09) |
| Display thickness | $X_4$ (mm) | (0.91, 1.09) |
| Lens thickness | $X_5$ (mm) | (0.91, 1.09) |
| Young modulus of mainboard | $P_1$ (MPa) | (10400, 11600) |
| Young modulus of Display | $P_2$ (MPa) | (22600, 23400) |
| Young modulus of Lens | $P_3$ (MPa) | (2380, 2580) |
| Power consumption of chip 1 | $P_4$ (W) | (0.09, 0.21) |
| Power consumption of chip 2 | $P_5$ (W) | (0.09, 0.21) |

The DES-PSO proposed in subsection 4.2 and the original HCLPSO are used to simultaneously optimize the six response surface functions listed in Table 3, with the computational results presented in Table 4. It can be seen from Table 4 that both algorithms can obtain the exact upper and lower bounds for the six surrogate models, indicating that both the improved HCLPSO algorithm and the original HCLPSO algorithm have good optimization accuracy. To further compare the computational efficiency of the two algorithms, the number of convergence steps required for each algorithm to find the exact solutions for the interval upper and lower bounds is counted and listed in Table 5.

From Table 5, it can be seen that DES-PSO has a significantly faster convergence rate than the original HCLPSO algorithm. In the process of determining the lower bounds of the six response surface functions, the convergence speed of DES-PSO is on average 57.96% faster than the original algorithm, with the highest improvement being 62.80%. In the process of determining the upper bounds of the six response surface functions, the convergence speed of the improved algorithm is on average 56.71% faster than the original algorithm, with the maximum improvement being 60.56%. These results indicate that DES-PSO can efficiently handle the considered uncertainty interval analysis problems.

Table 3. Six response surface functions.

| Variable | Response surface function |
|---|---|
| $\sigma_1^N$ (MPa) | $\sigma_1^N = \left(0.001848 P_2^2 - 0.3688 P_2 P_3 + 973.18 P_2 + 1.609 P_3^2\right) \times 10^{-6} - 30.19 X_1^2$ $+ 1.133 X_1 X_3 + 33.10 X_1 X_4 + 1.313 X_1 X_5 + 0.4128 X_3^2 - 3.7317 X_3 X_4$ $- 0.26871 X_3 X_5 - 56.55 X_4^2 + 65.54 X_4 X_5 - 55.32 X_5^2 + 129.86$ |
| $\sigma_2^N$ (MPa) | $\sigma_2^N = \left(-0.03509 P_2^2 + 0.1813 P_2 P_3 + 1277 P_2 - 1.461 P_3^2\right) \times 10^{-6} - 35.80 X_1^2$ $+ 6.112 X_1 X_3 + 32.86 X_1 X_4 + 2.891 X_1 X_5 - 6.809 X_3^2 + 4.303 X_3 X_4$ $+ 9.209 X_3 X_5 - 63.71 X_4^2 + 67.43 X_4 X_5 - 64.37 X_5^2 + 135.2$ |
| $\sigma_3^N$ (MPa) | $\sigma_3^N = \left(0.03054 P_2^2 - 0.95 P_2 P_3 + 802.6 P_2 + 4.645 P_3^2\right) \times 10^{-6} - 28.19 X_1^2$ $+ 4.188 X_1 X_3 + 28.63 X_1 X_4 + 0.2030 X_1 X_5 + 9.152 X_3^2 - 16.12 X_3 X_4$ $- 15.75 X_3 X_5 - 42.17 X_4^2 + 62.61 X_4 X_5 - 36.32 X_5^2 + 119.5$ |
| $\sigma^H$ (MPa) | $\sigma^H = 0.0000002578 P_1^2 - 0.00002501 P_1 X_2 - 0.9103 X_1^2 + 0.02502 X_1 X_2$ $+ 0.6950 X_1 X_3 + 0.1007 X_2^2 + 0.0125 X_2 X_3 - 2.372 X_3^2 + 37.54$ |
| $T_1$ (°C) | $T_1 = 0.5473 X_1^2 - 2.932 X_1 X_2 - 0.3207 X_1 X_3 + 5.589 X_2^2$ $- 2.970 X_2 X_3 - 1.206 X_3^2 + 71.85 P_4 + 72.81 P_5 + 299.3 P_4 P_5 + 62.05$ |
| $T_2$ (°C) | $T_2 = 0.5448 X_1^2 - 2.923 X_1 X_2 - 0.3219 X_1 X_3 + 5.569 X_2^2$ $- 2.973 X_2 X_3 - 1.204 X_3^2 + 61.10 P_4 + 96.78 P_5 + 255.2 P_4 P_5 + 61.11$ |

Table 4. Comparison of the results of two algorithms.

| Variable | Lower bound of the interval | | Upper bound of the interval | |
|---|---|---|---|---|
| | DES-PSO | HCLPSO | DES-PSO | HCLPSO |
| $\sigma_1^N$ (MPa) | 89.060 | 89.060 | 105.318 | 105.318 |
| $\sigma_2^N$ (MPa) | 66.908 | 66.908 | 89.221 | 89.221 |
| $\sigma_3^N$ (MPa) | 26.612 | 26.612 | 47.865 | 47.865 |
| $\sigma^H$ (MPa) | 62.230 | 62.230 | 69.937 | 69.937 |
| $T_1$ (°C) | 75.104 | 75.104 | 105.603 | 105.603 |
| $T_2$ (°C) | 74.984 | 74.984 | 105.475 | 105.475 |

Table 5. Comparison of convergence steps of two algorithms.

| Variable | Lower bound of the interval | | Upper bound of the interval | |
|---|---|---|---|---|
| | DES-PSO | HCLPSO | DES-PSO | HCLPSO |
| $\sigma_1^N$ | 958 | 2287 | 985 | 2249 |
| $\sigma_2^N$ | 1114 | 2499 | 1070 | 2436 |
| $\sigma_3^N$ | 1061 | 2415 | 1105 | 2480 |
| $\sigma^H$ | 667 | 1793 | 773 | 1960 |
| $T_1$ | 922 | 2224 | 1000 | 2313 |
| $T_2$ | 998 | 2314 | 1095 | 2446 |
| Average improvement | 57.96% | | 56.71% | |

## 6. Couclusions

In order to construct an accurate and efficient method for interval uncertainty analysis, this paper introduces HCLPSO into interval analysis. Aiming at the issue of high computational cost in traditional HCLPSO, this paper optimizes the random search mechanism of HCLPSO with the LDS generated by the DES, and constructs a new efficient interval analysis method, DES-PSO. By applying the newly proposed method to the solution of interval equation systems and the design of smartwatch casings, this paper demonstrates that the newly proposed DES-PSO can significantly improve computational efficiency while ensuring computational accuracy.

This paper only uses the DES-PSO algorithm to handle the solution of interval models, and in the future, we will also consider applying this precise and efficient DES-PSO method to interval processes and interval field problems [76, 77].

**Author contributions**

**Xuanlong Wu**: Methodology, Software, Validation, Writing – original draft, Writing – review & editing. **Peng Zhong**: Conceptualization, Methodology, Project administration, Supervision, Funding acquisition, Writing – review & editing. **Weihao Lin**: Methodology, Writing – review & editing, Resources.

**Use of AI tools declaration**

The authors declare they have not used Artificial Intelligence (AI) tools in the creation of this article.

**Conflict of interest**

We declare that we have no financial and personal relationships with other people or organizations that can inappropriately influence our work, there is no professional or other personal interest of any nature or kind in any product, service and/or company that could be construed as influencing the position presented in, or the review of, the manuscript entitled.